\def\year{2021}\relax
\documentclass[letterpaper]{article} 
\usepackage{aaai21}  
\usepackage{times}  
\usepackage{helvet} 
\usepackage{courier}  
\usepackage[hyphens]{url}  
\usepackage{graphicx} 
\usepackage{natbib}  
\usepackage{caption} 
\usepackage{amsthm,amsmath,amssymb}
\usepackage{mathrsfs}  
\usepackage{amssymb}
\usepackage{booktabs}
\usepackage{textcomp}
\usepackage{color,xcolor}
\usepackage[switch]{lineno}
\title{Structure-Consistent Weakly Supervised Salient Object Detection with Local Saliency Coherence }
\author{
    Siyue Yu\textsuperscript{\rm 1},
    Bingfeng Zhang\textsuperscript{\rm 1},
    Jimin Xiao\textsuperscript{\rm 1}\thanks{Corresponding Author.},
    Eng Gee Lim \textsuperscript{\rm 1} \\
    
}
\affiliations{
    \textsuperscript{\rm 1}School of Advanced Technology, Xi'an Jiaotong Liverpool University \\
    siyue.yu@xjtlu.edu.cn, bingfeng.zhang@xjtlu.edu.cn, jimin.xiao@xjtlu.edu.cn, enggee.lim@xjtlu.edu.cn 
}

\begin{document}
\maketitle

\begin{abstract}
Sparse labels have been attracting much attention in recent years. However, the performance gap between weakly supervised and fully supervised salient object detection methods is huge, and most previous weakly supervised works adopt complex training methods with many bells and whistles.
In this work, we propose a one-round end-to-end training approach for weakly supervised salient object detection via scribble annotations without pre/post-processing operations or extra supervision data. 
Since scribble labels fail to offer detailed salient regions, we propose a local coherence loss to propagate the labels to unlabeled regions based on  image features and pixel distance, so as to predict integral salient regions with complete object structures.
We design a saliency structure consistency loss as self-consistent mechanism to ensure consistent saliency maps are predicted with different scales of the same image as input, which could be viewed as a regularization technique to enhance the model generalization ability.
Additionally, we design an aggregation module (AGGM) to better integrate high-level features, low-level features and global context information for the decoder to aggregate various information.  
Extensive experiments show that our method achieves a new state-of-the-art performance on six benchmarks (e.g. for the ECSSD dataset: $F_\beta = 0.8995$, $E_\xi = 0.9079$ and \emph{MAE} $= 0.0489$), with an average gain of 4.60\% for F-measure, 2.05\% for E-measure and 1.88\% for MAE over the previous best method on this task. Source code is available at \url{http://github.com/siyueyu/SCWSSOD}.
\end{abstract}

\section{Introduction}
\begin{figure}
	\includegraphics[scale=0.34]{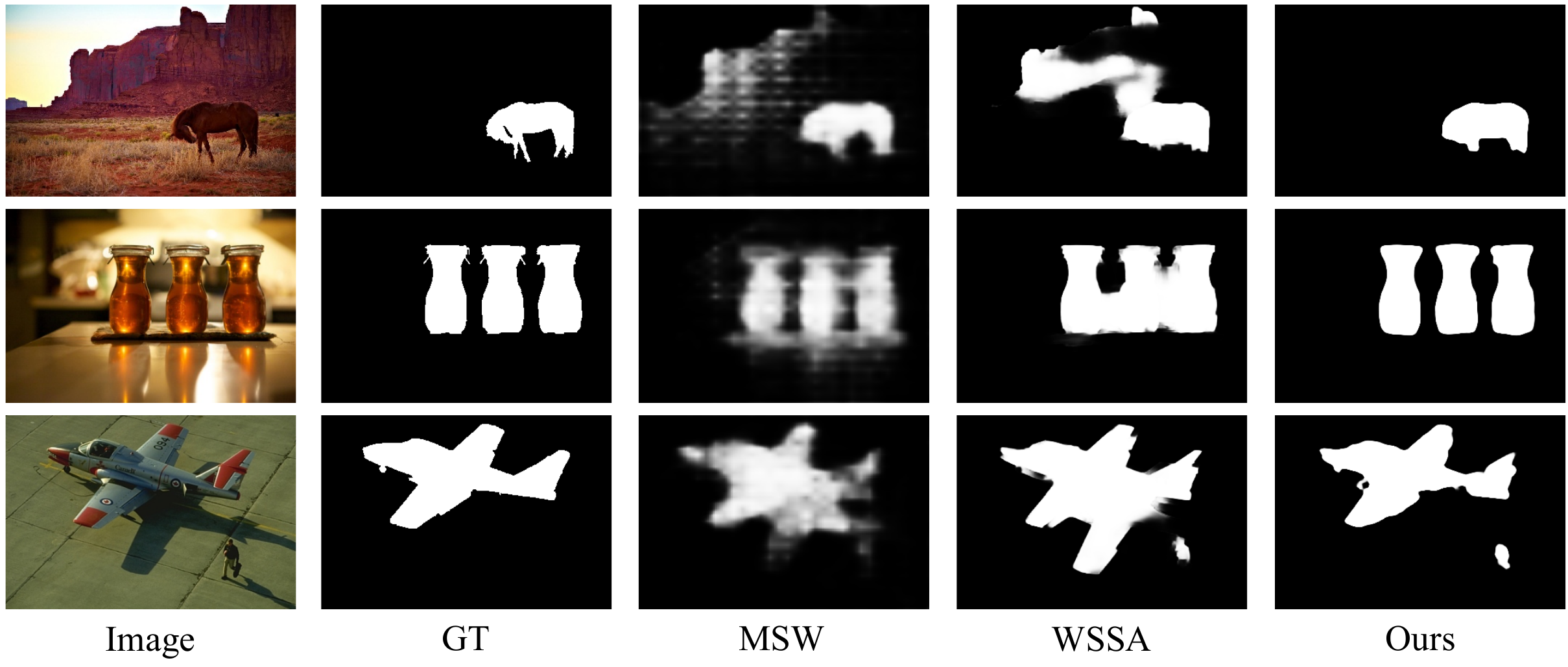}
	\caption{Our predicted saliency maps are compared with that of other weakly supervised methods. From left to right: Input image; Ground-truth; MSW~\cite{zeng2019multi}; WSSA~\cite{Zhang_2020_CVPR}; Ours.}
	\label{FIG:comparison1}
\end{figure}
Salient object detection (SOD) aims to detect the most attractive regions  in an image according to the human perception. It can be further applied in different computer vision tasks, such as image-sentence matching~\cite{Ji_2019_ICCV}, image segmentation~\cite{sun2019saliency} and image retrieval~\cite{wang2020visual}. In the last decade, deep learning based salient object detection algorithms~\cite{chen2020global, Liu_2019_CVPR, Qin_2019_CVPR} have
become popular due to their superior performance. These methods usually design different modules to help their networks learn better feature representations for saliency prediction. However, they are highly dependent on pixel-wise saliency labels, which are time-consuming and costly with manual annotations. 

In recent years, sparse labeling methods have attracted much attention. Many weakly supervised salient object detection methods have been proposed to improve label efficiency while maintaining model performance. Image level labels are utilized in some methods~\cite{wang2017learning, li2018weakly} to learn salient object detection.
However, these works usually use image-level tags for saliency localization and then further train their models with predicted saliency maps through multiple-stage learning. Besides, some other works ~\cite{NIPS2019_8314, Zhang_2017_ICCV} train their models with noisy pseudo labels from handcrafted methods and/or predicted maps by other weakly supervised SOD models, where pre-processing steps are used to clean noisy labels. All above mentioned works need complex training steps to obtain final saliency maps.

Additionally, scribble annotations are proposed recently due to their flexibility to label winding objects and low-cost compared to annotating per-pixel saliency masks.
However, scribble annotations cannot cover the whole object region or directly provide object structure. Therefore, edge detection is used in the framework~\cite{Zhang_2020_CVPR} to obtain object boundaries, and the SOD model is trained with predicted edge maps from other trained edge detection models. However, this step introduces extra data information into the SOD training process to recover integral object structure. The training process of \cite{Zhang_2020_CVPR} is also complex, as they design a scribble boosting scheme to iteratively train their model using initial saliency predictions to obtain higher quality saliency maps.

In this paper, we aim to tackle the aforementioned issues in existing weakly supervised SOD methods. Specifically, we aim to design a high performance SOD method with scribble annotations via one-stage end-to-end training, where no pre/post-processing steps nor extra supervision (e.g., edge maps) will be used. To mitigate the issue of poor boundary localization caused by scribble annotations and partial cross-entropy loss~\cite{Zhang_2020_CVPR}, we design a local saliency coherence loss to provide supervision for unlabeled points, based on the idea that points with similar features and/or close positions should have similar saliency values. By doing this, we take advantage of intrinsic properties of an image instead of extra edge or other assisting information to help our model learn better object structure and predict integral salient regions. 

Besides, we find that weakly supervised SOD models fail to predict consistent saliency maps with different scales of the same image as input. To handle this problem, 
we propose a saliency structure consistency loss, which could be viewed as a regularization technique to enhance the model generalization ability.

Additionally, global context information can infer the relationship of different salient regions and help network predict better results~\cite{chen2020global}. High-level features can provide better semantic information and low-level features can capture rich spatial information~\cite{hou2017deeply}. In the decoder layers, we design an aggregation module called AGGM to integrate all information for better feature representations using the attention mechanism~\cite{shi2020weakly}. 

With our specially designed loss functions and network structures, our model can predict saliency maps close to human perception. Some obtained saliency maps are illustrated in Fig.~\ref{FIG:comparison1}. Our method predicts smoother and integral saliency objects even for the challenging cases with background disturbance, object shadow, and multiple objects. 
In general, our main contributions can be summarized as:

\begin{itemize}
	\item A local saliency coherence loss is proposed for scribble supervised saliency object detection, which helps our network to learn integral salient object structures without any extra assisting data or complex training process. 
	
	\item A self-consistent mechanism is introduced to ensure that consistent saliency masks will be predicted with different scales of the same image as input. It is an effective regularization  to enhance the model generalization ability.
	
	\item An aggregation module named AGGM is designed in the encoder-decoder framework for weakly supervised SOD, which effectively aggregates global context information as well as high-level and low-level features. 
	
	\item  Comprehensive experiments show that our approach achieves a new state-of-the-art performance compared with other scribble supervised SOD algorithms on six widely-used benchmarks, with an average gain of 4.60\% for F-measure, 2.05\% for E-measure and 1.88\% for MAE over the previous best performing method. 
\end{itemize}

\begin{figure*}
	\centering
	\includegraphics[scale=0.63]{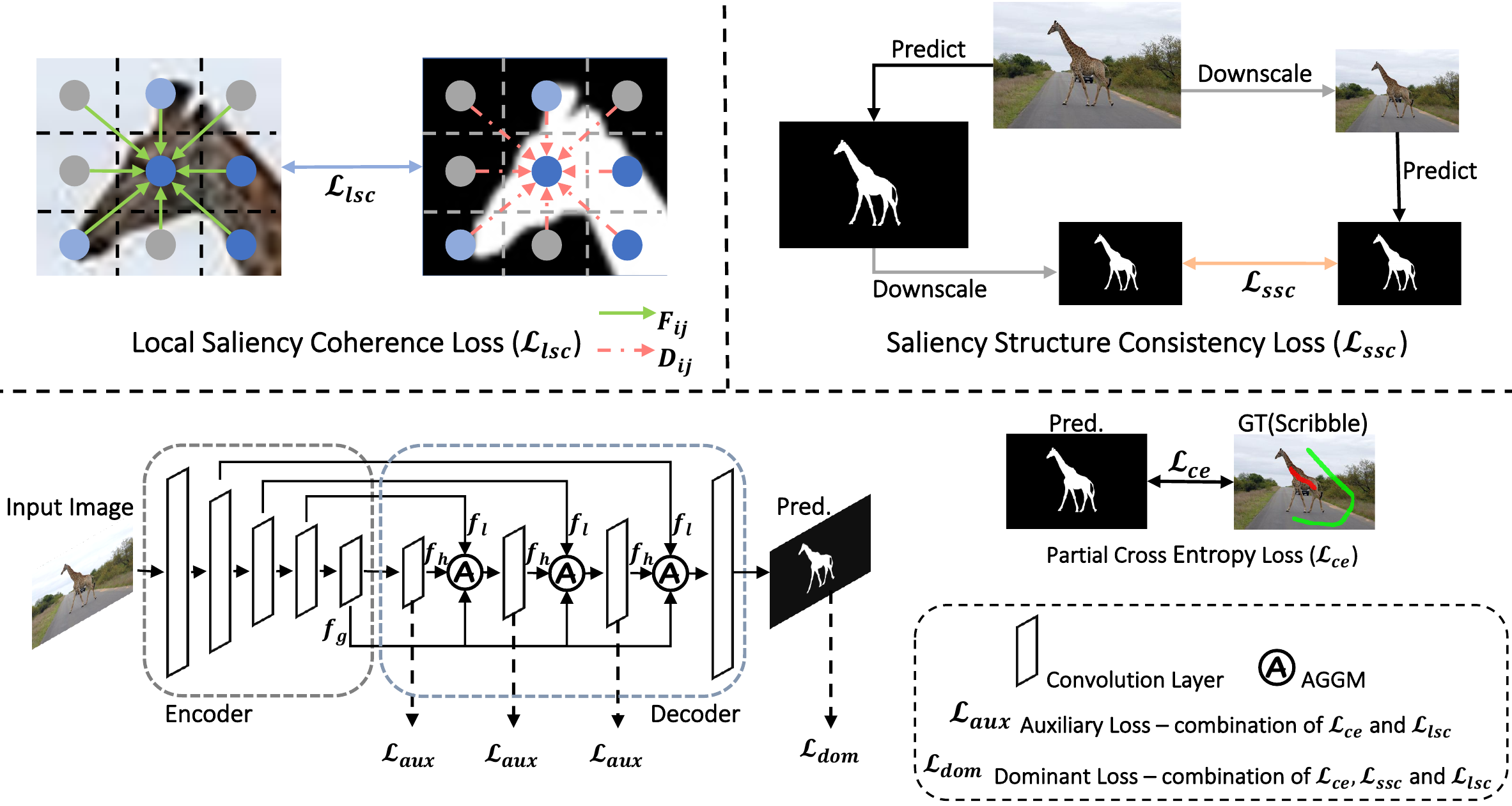}
	\caption{The framework of our network and learning procedure. Specifically, $f_l$, $f_h$, $f_g$ denote to the low-level, high-level features and global context information, respectively. The AGGM is applied in the decoder to integrate multi-level features. The proposed local saliency coherence loss and saliency structure consistency loss are applied with partial cross entropy loss to optimize the network as a dominant loss. To facilitate optimization, our local saliency coherence loss are applied with partial cross entropy loss as auxiliary losses to further supervise intermediate low-resolution saliency maps.}
	\label{FIG:overview}
\end{figure*}

\section{Related Work}
\subsubsection{Salient Object Detection.}
CNN based salient object detection is popular due to its superior performance under the fully supervised setting. Most proposed methods~\cite{chen2020global,zhang2018progressive,Pang_2020_CVPR} studied how to aggregate multi-level features in the network to better capture features for salient regions. Some works~\cite{Qin_2019_CVPR,Wang_2019_CVPR,Zhao_2019_ICCV} introduced edge supervision to learn object boundary and in turn refine their saliency predictions with better object structures. However, these methods need fully pixel-level annotations, which are costly and time-consuming. Thus, in this paper, we focus on weakly supervised salient object detection.

\subsubsection{Weakly Supervised Salient Object Detection.}
With recent advances in weakly supervised learning, SOD considers weakly supervised or unsupervised learning to ease the burden on pixel level annotations. WSS~\cite{wang2017learning} firstly proposed to learn SOD through image-level labels to reduce annotation efforts. They proposed a global smooth pooling layer and a foreground inference network to maintain prediction quality for unseen categories during inference. Additionally, MSW~\cite{zeng2019multi} utilized diverse supervision sources including category labels, captions and noisy pseudo labels to train their network via their attention transfer loss. Besides, user-friendly scribble annotations was proposed in ~\cite{Zhang_2020_CVPR}, which took only 1$\sim$2 seconds to label one image. 
To simplify training procedures and narrow the gap between weakly and fully supervised methods, we propose an end-to-end training approach containing one-round training without post-processing.

\subsubsection{Weakly Supervised Semantic Segmentation.}
Various types of weak annotations were proposed to save human labor expense for semantic segmentation. Bounding boxes were used in BCM~\cite{Song_2019_CVPR} to provide the regions of foreground objects. Khoreva et al~\cite{khoreva2017simple} explored bounding box supvervison for both instance and semantic segmentation. Berman et al~\cite{bearman2016s} proposed point-level supervision. DEXTR~\cite{Maninis_2018_CVPR} made the use of both bounding box and point seed label to predict semantic segmentation. Additionally, AE~\cite{chen2018tap} presented a new interactive segmentation by balancing the selection and defocus cues at phototaking based on point-level supervision. RRM~\cite{zhang2019reliability} proposed a fully end-to-end training method to learn to predict segmentation maps through image-level labels. Additionally, SEAM~\cite{wang2020self} designed a self-supervised equivariant attention mechanism to supervise CAM with image-level labels. ScrribleSup~\cite{Lin_2016_CVPR} used a graphical model to learn from scribbles. Besides, Normalized cut loss~\cite{Tang_2018_CVPR} and kernel cut loss~\cite{Tang_2018_ECCV} were proposed to regularize cross entropy loss for semantic segmentation with scribble annotations. Although scribble annotations for semantic segmentation have been intensively studied, little effort was devoted on SOD with scribble labels. 

\section{Methodology}

\subsection{Overview}
Firstly, the training dataset is defined as $U=\{x_n, y_n\}_{n=1}^N$, where $x_n$ is the input image, $y_n$ is the corresponding label,  and $N$ is the total number of training images. Note that, in our task, the label $y_n$ is annotated as scribble. 

The whole network and learning framework are shown in Fig.~\ref{FIG:overview}. The network contains an encoder and a decoder, and the designed aggregation module AGGM is applied in each layer of the decoder. In this way, the three kinds of information can be better propagated into next layers. Sigmoid function is applied on the output of the decoder to normalize the output saliency values to $[0, 1]$. Thus, our network receives the input image and outputs the corresponding saliency map directly. During training, the proposed local saliency coherence loss and saliency structure consistency loss are applied with the partial cross entropy loss as the dominant loss to supervise the final predicted saliency map. Meanwhile, to facilitate training, we enforce an auxiliary loss, which includes the local saliency coherence loss and the partial cross entropy loss on each sub-stage to supervise the intermediate low-resolution saliency maps. Note that the intermediate low-resolution saliency maps are upsampled to the corresponding scale of $x_n$ to supervise the intermediate low-resolution saliency map. In the following sections, we will discuss the AGGM and each loss in details.

\subsection{Aggregation Module}
\begin{figure}
	\centering
	\includegraphics[scale=0.4]{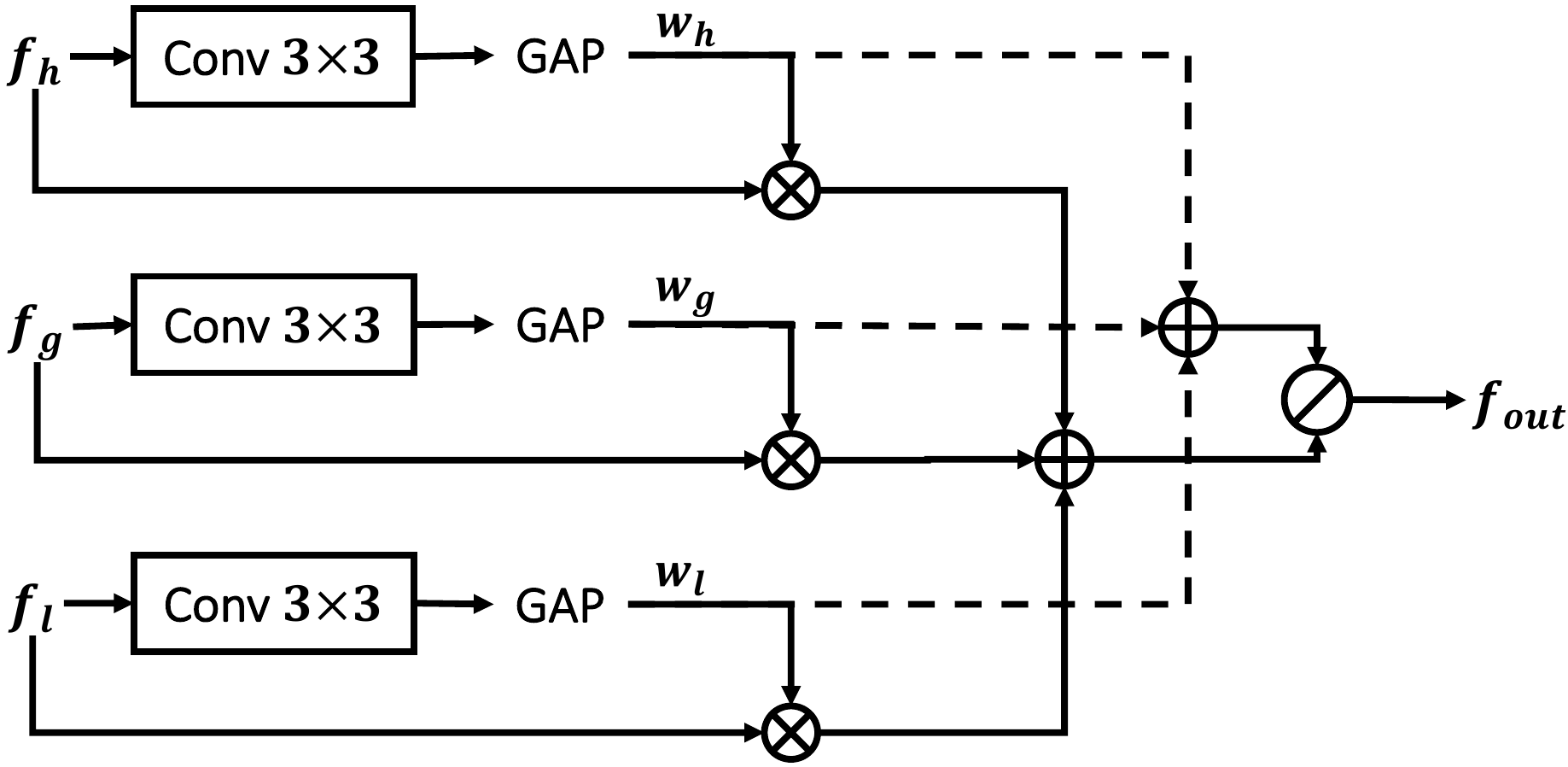}
	\caption{Framework of AGGM, where `GAP' denotes to global average pooling, `$\times$' is multiplication, `$+$' is addition and `$/$' is division.}
	\label{FIG:AGGM}
\end{figure}

Our network follows an encoder-decoder framework. The encoder layers learn different features of the salient regions and further propagate them to the decoder.
Since some detailed features might be diluted, each decoding layer uses output of preceding layer, the features of corresponding encoding layer and the global context information, which is the final output of the encoder, as inputs to predict salient regions. However, we argue that each input should be allocated to different weights for each decoding layer. To learn the importance of each input feature by self-learning, our aggregation module is designed as in Fig.~\ref{FIG:AGGM}. $3\times 3$ convolution layers and global average pooling layers are applied to learn the importance of each input feature. Once the weights are obtained, normalization is applied, which can be written as: 
\begin{equation}
\centering
f_{out} = \frac{w_{h}f_{h} + w_{g}f_{g} + w_{l}f_{l}}{w_{h} + w_{g}+ w_{l}},
\label{eq:aggm}
\end{equation}
where $w_{h}$, $w_{g}$ and $w_{l}$ are obtained weights. $f_{h}$, $f_{g}$ and $f_{l}$ are input multi-level features. With Eq.~(\ref{eq:aggm}), the network can aggregate multi-level features and with our AGGM, the importance of each feature can be learned by self-learning.

\subsection{Local Saliency Coherence Loss}
For scribble annotations, there are a great number of unlabeled pixels. With only the given scribble labels, it is hard to learn rich information of salient regions. In addition, there is no category information in the SOD task, making it more difficult to learn object structures. Therefore, the network needs other supervision to get better saliency maps with clear object boundaries. In this case, we design a local saliency coherence loss to help network predict smooth saliency maps with the scribble annotations. We consider that for pixel $i$ and pixel $j$ of the same input image, if they are with similar features or close positions, they tend to have similar saliency scores. On the other hand, if two points do not share similar features or they are distant from each other, they are more likely to have different saliency scores. We first define the saliency difference between two different pixels $i$ and $j$ as follows: 
\begin{equation}
\centering
D(i, j) = \lvert S_{i} - S_{j} \rvert,
\label{eq:saliency difference}
\end{equation}
where $S_{i}$ and $S_{j}$ are the predicted saliency scores of pixels $i$ and $j$, respectively. We use $L1$ distance to directly compute the discrepancy of the saliency scores.

Instead of computing the similarities between any two pixels in an image, which introduces too much background noise and takes too much GPU memory, we compute the discrepancy of a reference point with its adjacent points in a $k \times k$ kernel size. However, if we directly compute the loss using Eq.~(\ref{eq:saliency difference}), the network fails to distinguish the salient region with background, especially for the pixels close to the boundaries. For pixels around object boundaries, their saliency scores are not always similar with their adjacent pixels. Therefore, we set a similarity energy between two pixels $i$ and $j$, which is defined based on Gaussian kernel bandwidth filter~\cite{obukhov2019gated}, to draw close saliency scores for pixels with similar features and/or with small distance. Then, the final local saliency coherence loss is designed as:
\begin{equation}
\centering
\mathcal{L}_{lsc} = \sum_{i} \sum_{j \in K_i} F(i, j)D(i, j), 
\label{eq:LSC Loss}
\end{equation}
where $K_i$ is the region covered by a  $k \times k$ kernel around pixel $i$, and $F(i, j)$ denotes to the following filter:
\begin{equation}
\centering 
F(i, j) = \frac{1}{w}\exp(-\frac{{\|P(i) - P(j)}\|^2}{2\sigma_{P}^2}-\frac{\|I(i) - I(j)\|^2}{2\sigma_{I}^2}),
\label{eq:pairwise_potentials}
\end{equation}
where $1/w$ is the normalized weight, $P(\cdot)$ and $I(\cdot)$ are the position and RGB color of a pixel, respectively. $\sigma_{P}$ and $\sigma_{I}$ are hyper parameters for the scale of Gaussian kernels. $\|\cdot \|^2$ is an $L2$ operation.

The local saliency coherence loss $\mathcal{L}_{lsc}$ enforces similar pixels in the kernel to share consistent saliency scores, which further propagates labeled points to the whole image during training. With the partial cross entropy loss to supervise labeled points, the network can acquire enlarged salient regions with limited labels without any extra information. 

\subsection{Self-consistent Mechanism}
For a good salient object detection model, saliency maps predicted with different scales of the same image should be consistent. 
We define a salient object detection function as $f_{\theta}(\cdot)$ with parameter $\theta$, and a transformation as $T(\cdot)$. Then, for an ideal $f_{\theta}(x)$, it should satisfy this equation: 
\begin{equation}
\centering
f_{\theta}(T(x)) = T(f_{\theta}(x)).
\label{eq:transform}
\end{equation}
\begin{figure}
	\centering
	\includegraphics[scale=0.45]{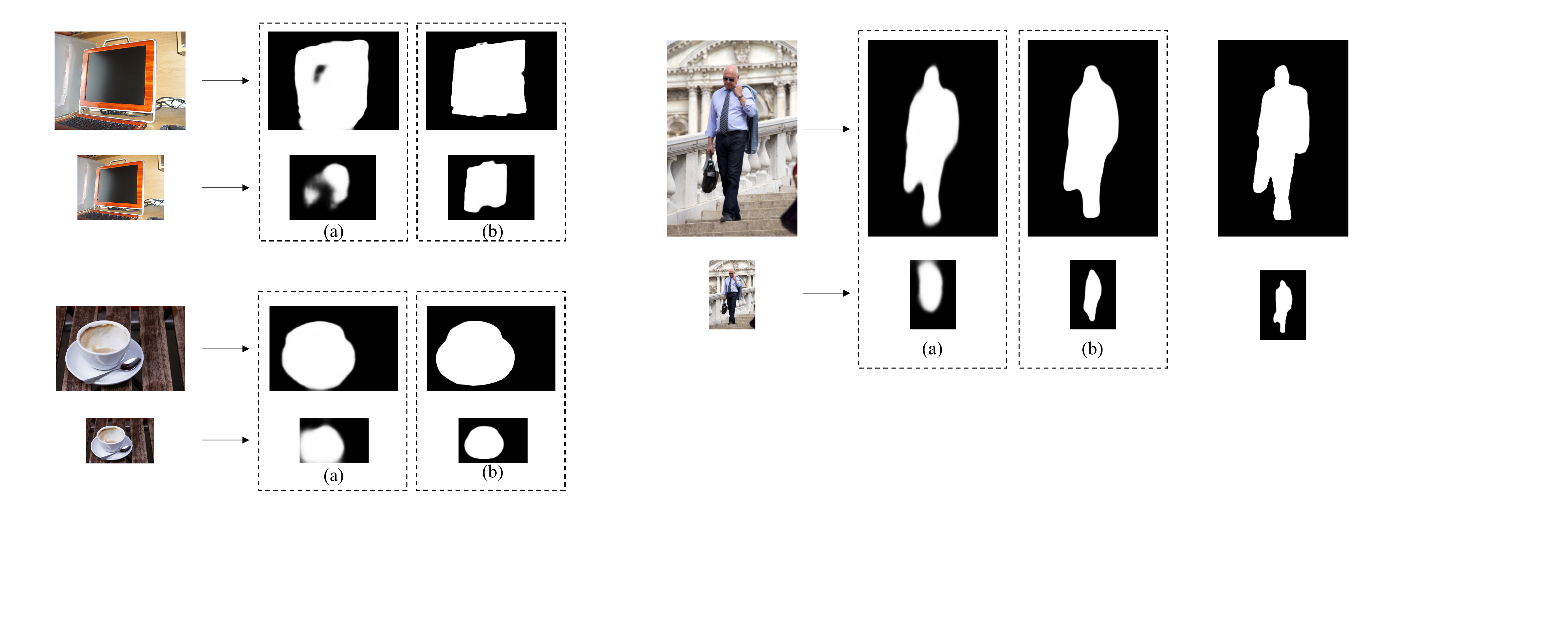}
	\caption{Comparison of predicted saliency maps for an input image with different scales: (a) without self-consistent mechanism; (b) with self-consistent mechanism.}
	\label{FIG:scale}
\end{figure}
However, we find that it is difficult for weakly supervised SOD networks to predict consistent saliency maps with different input scales, as shown in Fig.~\ref{FIG:scale}(a). Therefore, by considering Eq.~(\ref{eq:transform}) as a regularization, we design a structure consistency loss on predicted saliency maps from different input scales, which is defined as follows: 
\begin{equation}
\centering
\mathcal{L}_{ssc} = \frac{1}{M} \sum_{u,v}\alpha \frac{1-SSIM(S_{u,v}^\Downarrow, S_{u,v}^\downarrow)}{2}+(1-\alpha)\lvert S_{u,v}^\Downarrow- S_{u,v}^\downarrow\rvert,
\label{eq:structure_consistency_loss}
\end{equation}
where $S^\downarrow$ is down-scaled predicted saliency map of a normal input image, $S^\Downarrow$ is the predicted saliency map of the same image with down-scaled size, and $M$ is the number of pixels. SSIM denotes to the single scale SSIM~\cite{wang2004image, godard2017unsupervised} and $\alpha = 0.85$~\cite{godard2017unsupervised}. With Eq.~(\ref{eq:structure_consistency_loss}), the network can learn more information on object structure and enhance generalization ability for different input scales. As shown in Fig.~\ref{FIG:scale}(b), with our self-consistent mechanism, the network can adapt to different scales and predict saliency maps with better object structure. 

\subsection{Objective Function}
As shown in Fig.~\ref{FIG:overview}, our final loss function is the combination of a dominant loss and auxiliary losses following GCPANet~\cite{chen2020global}. Specifically, a $3 \times 3$ convolution layer is conducted to squeeze the channel to 1 at each stage of the decoder to compute the saliency scores. Then, our proposed losses are used to cooperate with the partial cross entropy loss, which can be written as:
\begin{equation}
\centering
\mathcal{L}_{ce} =\sum_{i \in \tilde{\mathcal{Y}}} -y_ilog\hat{y}_i - (1-y_i)log(1-\hat{y}_i),
\label{eq:CE}
\end{equation}
where $y$ denotes for ground-truth, $\hat{y}$ is the predicted values and $\tilde{\mathcal{Y}}$ is the set of labeled pixels via scribble annotations.

The auxiliary loss $\mathcal{L}_{aux} $ and dominant loss $\mathcal{L}_{dom}$ can be written as:
\begin{equation}
\centering
\mathcal{L}_{aux}^q = \begin{array}{cc}
\mathcal{L}_{ce} + \beta \mathcal{L}_{lsc} & q \in \left\{1, 2, 3\right\}\end{array},
\label{eq:aux}
\end{equation}
\begin{equation}
\centering
\mathcal{L}_{dom} = \mathcal{L}_{ce} + \mathcal{L}_{ssc} + \beta \mathcal{L}_{lsc},
\label{eq:dom}
\end{equation}
where the hyper-parameter $\beta$ shares the same value in Eq.~(\ref{eq:aux}) and Eq.~(\ref{eq:dom}), and $q$ in Eq.~(\ref{eq:aux}) stands for the index of a decoder layer.

Finally, the overall objective function of our network is:
\begin{equation}
\centering
\mathcal{L}_{total} = \mathcal{L}_{dom} + \sum_{q = 1}^3 \lambda _q \mathcal{L}_{aux}^q,
\label{eq:objective function}
\end{equation}
where $\lambda_q$ is to balance the auxiliary loss of each stage, and we take the same value as in GCPANet~\cite{chen2020global}.

\section{Experiments}
\subsection{Implementation Details and Setup}
\subsubsection{Implementation Details.}
We use GCPANet~\cite{chen2020global} with backbone of ResNet-50~\cite{he2016deep} pretrained on ImageNet~\cite{deng2009imagenet} as baseline. The partial cross entropy loss (Eq.~(\ref{eq:CE})) is computed for background and foreground individually. $w$, $\sigma_{P}$, and $\sigma_{I}$ in Eq.~(\ref{eq:pairwise_potentials}) are set to 1, 6 and 0.1, respectively. $\beta$ in Eq.~(\ref{eq:aux}) and Eq.~(\ref{eq:dom}) is set to 0.3.
The model is optimized by SGD with batch size of 16, momentum of 0.9 and weight decay of $\text{5} \times \text{10}^{\text{-4}}$. Additionally, we use triangular warm-up and decay strategies with the maximum learning rate of 0.01 and the minimum learning rate of $\text{1} \times \text{10}^{\text{-5}}$ to train the network with 40 epochs. During training, each image is resized to 320$\times $320 with random horizontal flipping and random cropping. In the inference stage, input images are simply resized to $320\times 320$ and then fed into the network to predict saliency maps without any post-processing. All experiments are run on NVIDIA GeForce RTX 2080 Ti. 

\subsubsection{Datasets.}
We train our network on scribble annotated dataset S-DUTS~\cite{Zhang_2020_CVPR} and evaluate our model on six widely-used salient object detection benchmarks: (1) ECSSD~\cite{yan2013hierarchical}; (2) DUT-OMRON~\cite{yang2013saliency}; (3) PASCAL-S~\cite{li2014secrets}; (4) HKU-IS~\cite{li2015visual}; (5) THUR~\cite{cheng2014salientshape}; (6) DUTS-TEST~\cite{wang2017learning}.

\subsubsection{Baseline Methods and Evaluation Metrics.}
Our model is compared with 6 state-of-the-art weakly supervised or unsupervised SOD methods and 10 fully supervised SOD methods as baselines. We take three widely-used evaluation metrics for fair comparison: mean F-measure ($F_\beta$), mean E-measure ($E_\xi$)~\cite{fan2018enhanced}, and Mean Absolute Error (\emph{MAE})~\cite{cong2017iterative}.

\subsection{Comparison with State-of-the-arts}
\subsubsection{Quantitative Comparison.}
In Table~\ref{tbl:overall}, we compare our approach with other state-of-the-art approaches. It can be seen that our method achieves a new state-of-the-art performance among weakly supervised or unsupervised approaches under all the evaluation metrics. Our one-round training method obtains an average gain of 4.60\% for $F_\beta$, 2.05\% for $E_\xi$, and 1.88\% for \emph{MAE}, compared with the previous best two-round training method WSSA~\cite{Zhang_2020_CVPR} on the 6 datasets. Besides, our approach is comparable or even superior to some fully supervised methods, like PiCANet~\cite{liu2018picanet}, PAGR\cite{zhang2018progressive} and MLMSNet~\cite{wu2019mutual}.

\begin{table*}
	\caption{Comparison with other state-of-the-art approaches on 6 benchmarks. $\uparrow$ means that larger is better and $\downarrow$ denotes that smaller is better. The best performance on each dataset is highlighted in \textbf{boldface} under different cases of supervision. `Sup.' denotes for supervision information. `F' means fully supervised. `I' means image-level supervised. `S' means scribble-level supervised. `M' means multi-source supervised and `Un' is for unsupervised. `$\dagger$' means two-round training.}\label{tbl:overall}
	\centering
	\begin{tabular}{c|c|ccc|ccc|ccc}
		\bottomrule
		& & \multicolumn{3}{| c |}{ECSSD} &\multicolumn{3}{| c |}{DUT-OMRON}&\multicolumn{3}{| c }{PASCAL-S} \\
		Methods & Sup. & $F_\beta \uparrow$ & $E_\xi \uparrow$ & $MAE\downarrow$ & $F_\beta \uparrow$ & $E_\xi \uparrow$ & $MAE\downarrow$ & $F_\beta \uparrow$ & $E_\xi \uparrow$ & $MAE\downarrow$ \\
		\midrule
		DGRL~\shortcite{wang2018detect} & F & 0.9027 & 0.9371 & 0.043 & 0.7264 & 0.8446 & 0.0632 & 0.8289 & 0.8353 & 0.1150 \\
		PiCANet~\shortcite{liu2018picanet} & F & 0.8864 & 0.9128 & 0.0464 & 0.7173 & 0.8407 & 0.0653 & 0.7979 & 0.8330 & 0.0750 \\
		PAGR\shortcite{zhang2018progressive} & F & 0.8718 & 0.8869 & 0.0644 & 0.6754 & 0.7717 & 0.0709 & 0.7656 & 0.7545 & 0.1516 \\
		MLMSNet~\shortcite{wu2019mutual} & F & 0.8856 & 0.9218 & 0.0479 & 0.7095 & 0.8306 & 0.0636 & 0.8129 & 0.8219 & 0.1193 \\
		CPD~\shortcite{wu2019cascaded} & F & 0.917 & 0.925 & 0.037 & 0.747 & 0.866 & 0.056 & 0.824 & 0.849 & 0.072 \\
		AFNet~\shortcite{feng2019attentive} & F & 0.9008 & 0.9294 & 0.0450 & 0.7425 & 0.8456 & 0.0574 & 0.8241 & 0.8269 & 0.1155 \\
		PFAN~\shortcite{zhao2019pyramid} & F & 0.8592 & 0.8636 & 0.0467 & 0.7009 & 0.7990 & 0.0615 & 0.7544 & 0.7464 & 0.1372 \\
		BASNet~\shortcite{Qin_2019_CVPR} & F & 0.880 & 0.916 & 0.037 & 0.756 & 0.869 & 0.056 & 0.775 & 0.832 & 0.076 \\
		GCPANet~\shortcite{chen2020global} & F & 0.9184 & 0.927 & 0.035 & 0.7479 & 0.839 & 0.056 & 0.8335 & 0.861 & \textbf{0.061} \\
		MINet~\shortcite{Pang_2020_CVPR} & F & \textbf{0.924} & \textbf{0.953} & \textbf{0.033} & \textbf{0.756} & \textbf{0.873} & \textbf{0.055} & \textbf{0.842} & \textbf{0.899} &  0.064 \\
		\midrule
		SVF~\shortcite{Zhang_2017_ICCV} & Un & 0.7823 & 0.8354 & 0.0955 & 0.6120 & 0.7633 & 0.1076 & 0.7351 & 0.7459 & 0.1669 \\
		MNL~\shortcite{zhang2018deep} & Un & 0.8098 & 0.8357 & 0.0902 & 0.5966 & 0.7124 & 0.1028 & 0.7476 & 0.7408 & 0.1576 \\
		ASMO~\shortcite{li2018weakly} & I & 0.7621 & 0.7921 & 0.0681 & 0.6408 & 0.7605 & 0.0999 & 0.6532 & 0.6474 & 0.2055 \\
		WSS~\shortcite{wang2017learning} & I & 0.7672 & 0.7693 & 0.1081 & 0.5895 & 0.7292 & 0.1102 & 0.6975 & 0.6904 & 0.1843 \\
		MSW~\shortcite{zeng2019multi} & M & 0.7606 & 0.7876 & 0.0980 & 0.5970 & 0.7283 & 0.1087 & 0.6850 & 0.6932 & 0.1780 \\
		WSSA~\shortcite{Zhang_2020_CVPR} & S & 0.845 & 0.898 & 0.068 & 0.679 & 0.823 & 0.074 & 0.772 & 0.791 & 0.145 \\
		WSSA$\dagger$~\shortcite{Zhang_2020_CVPR} & S & 0.8650 & 0.9077 & 0.0610 & 0.7015 & 0.8345 & 0.0684 & 0.7884 & 0.7975 & 0.1399 \\
		Ours & S & \textbf{0.8995} & \textbf{0.9079} & \textbf{0.0489} & \textbf{0.7580} & \textbf{0.8624} & \textbf{0.0602} & \textbf{0.8230} & \textbf{0.8465} & \textbf{0.0779} \\
		\bottomrule
		\bottomrule
		& & \multicolumn{3}{| c |}{HKU-IS}&\multicolumn{3}{| c |}{THUR}&\multicolumn{3}{| c }{ DUTS-TEST} \\
		Methods & Sup. & $F_\beta \uparrow$ & $E_\xi \uparrow$ & $MAE\downarrow$ & $F_\beta \uparrow$ & $E_\xi \uparrow$ & $MAE\downarrow$ & $F_\beta \uparrow$ & $E_\xi \uparrow$ & $MAE\downarrow$ \\
		\midrule
		DGRL~\shortcite{wang2018detect} & F & 0.8844 & 0.9388 & 0.0374 & 0.7271 & 0.8378 & 0.0774 & 0.7989 & 0.8873 & 0.0512 \\
		PiCANet~\shortcite{liu2018picanet} & F & 0.8704 & 0.9355 & 0.0433 & - & - & - & 0.7589 & 0.8616 & 0.0506 \\
		PAGR\shortcite{zhang2018progressive} & F & 0.8638 & 0.8979 & 0.0475 & \textbf{0.7395} & \textbf{0.8417} & \textbf{0.0704} & 0.7781 & 0.8422 & 0.0555 \\
		MLMSNet~\shortcite{wu2019mutual} & F & 0.8780 & 0.9304 & 0.0387 & 0.7177 & 0.8288 & 0.0794 & 0.7917 & 0.8829 & 0.0490 \\
		CPD~\shortcite{wu2019cascaded} & F & 0.891 & 0.944 & 0.034 & - & - & - & 0.805 & 0.886 & 0.043 \\
		AFNet~\shortcite{feng2019attentive} & F & 0.8877 & 0.9344 & 0.0358 & 0.7327 & 0.8398 & 0.0724 & 0.8123 & 0.8928 & 0.0457 \\
		PFAN~\shortcite{zhao2019pyramid} & F & 0.8717 & 0.8982 & 0.0424 & 0.6833 & 0.8038 & 0.0939 & 0.7648 & 0.8301 & 0.0609  \\
		BASNet~\shortcite{Qin_2019_CVPR} & F & 0.895 & 0.946 & 0.032 & 0.7366 & 0.8408 & 0.0734 & 0.791 & 0.884 & 0.048 \\
		GCPANet~\shortcite{chen2020global} & F & 0.8984 & 0.920 & 0.031 & - & - & - & 0.8170 & 0.891 & 0.038\\
		MINet~\shortcite{Pang_2020_CVPR} & F & \textbf{0.908} & \textbf{0.961} & \textbf{0.028} & - & - & - & \textbf{0.828} & \textbf{0.917} & \textbf{0.037} \\
		\midrule
		SVF~\shortcite{Zhang_2017_ICCV} & Un & 0.7825 & 0.8549 & 0.0753 & 0.6269 & 0.7699 & 0.1071 & 0.6223 & 0.7629 & 0.1069 \\
		MNL~\shortcite{zhang2018deep} & Un & 0.8196 & 0.8579 & 0.0650 & 0.6911 & 0.8073 & 0.0860 & 0.7249 & 0.8525 & 0.0749\\
		ASMO~\shortcite{li2018weakly} & I & 0.7625 & 0.7995 & 0.0885 & - & - & - & 0.5687 & 0.6900 & 0.1156 \\
		WSS~\shortcite{wang2017learning} & I & 0.7734 & 0.8185 & 0.0787 & 0.6526 & 0.7747 & 0.0966 & 0.6330 & 0.8061 & 0.1000 \\
		MSW~\shortcite{zeng2019multi} & M & 0.7337 & 0.7862 & 0.0843 & - & - & - & 0.6479 & 0.7419 & 0.0912 \\
		WSSA~\shortcite{Zhang_2020_CVPR} & S & 0.835 & 0.911 & 0.055 & 0.696 & 0.824 & 0.085 & 0.728 & 0.857 & 0.068 \\
		WSSA$\dagger$~\shortcite{Zhang_2020_CVPR} & S & 0.8576 & 0.9232 & 0.0470 & 0.7181 & 0.8367 & 0.0772 & 0.7467 & 0.8649 & 0.0622\\
		Ours & S & \textbf{0.8962} & \textbf{0.9376} & \textbf{0.0375} & \textbf{0.7545} & \textbf{0.8430} & \textbf{0.0693} & \textbf{0.8226} & \textbf{0.8904} & \textbf{0.0487} \\
		\bottomrule
	\end{tabular}
\end{table*}

\begin{figure*}
	\centering
	\includegraphics[scale=0.4]{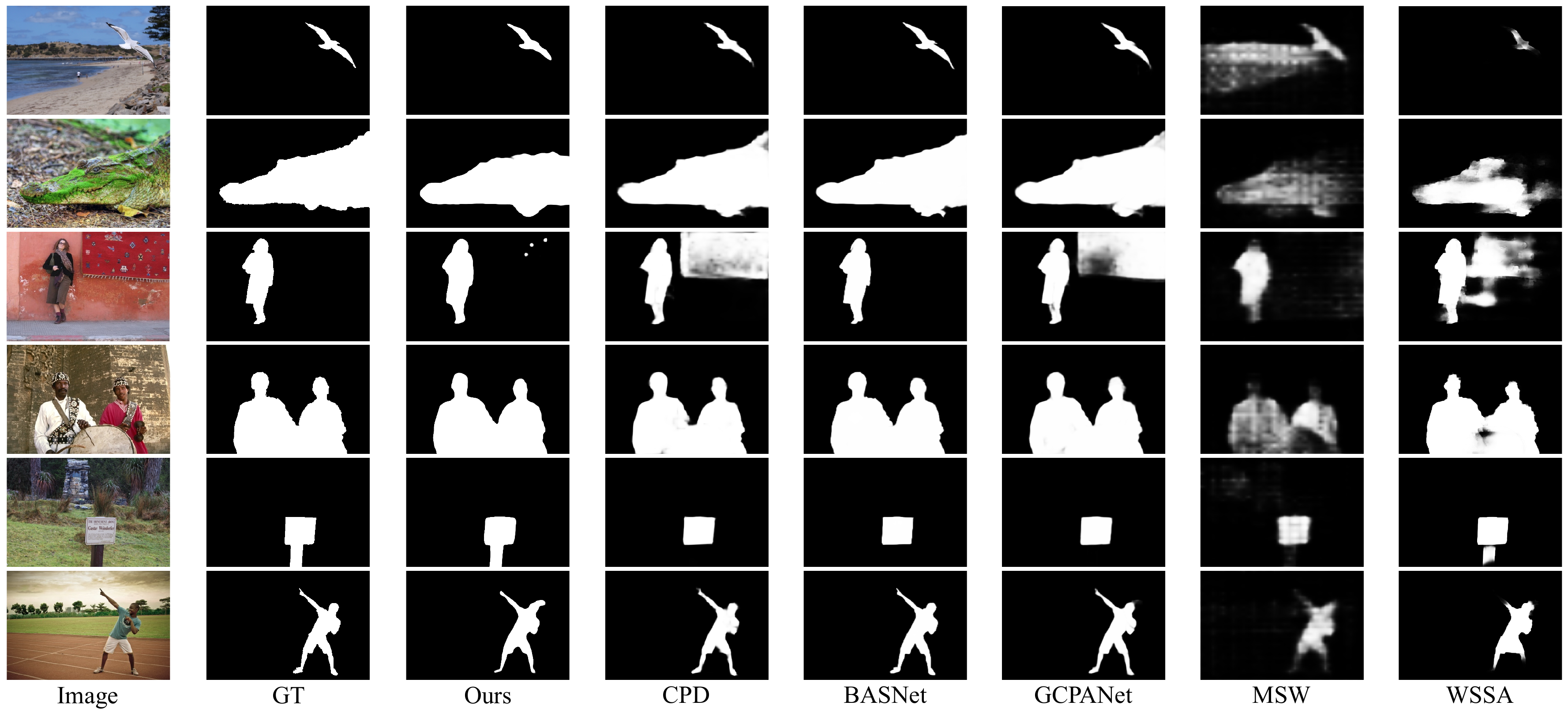}
	\caption{Qualitative comparisons of saliency maps predicted by our method and other state-of-the-art methods. Obviously, the maps predicted by ours are closer to the ground-truth compared with other weakly supervised approaches (MSW~\cite{zeng2019multi} \& WSSA~\cite{Zhang_2020_CVPR}), and some of our results even cover more details than that of fully supervised approaches (CPD~\cite{wu2019cascaded}, BASNet~\cite{Qin_2019_CVPR}, GCPANet~\cite{chen2020global}) as in row 5.}
	\label{FIG:overallcomparison}
\end{figure*}
\subsubsection{Qualitative Evaluation.}
We demonstrate some samples of our predicted saliency maps from the ECSSD dataset~\cite{yan2013hierarchical} in Fig.~\ref{FIG:overallcomparison}. It can be seen that our predicted saliency maps are more complete and precise compared with previous state-of-the-art weakly supervised methods (MSW and WSSA). Moreover, our approach is more general to different object classes and it is more robust to the disturbance of foreground-background (see rows 3 \& 4 in Fig.~\ref{FIG:overallcomparison}). In some cases, our approach even performs better than fully supervised methods, such as CPD, BASNet and GCPANet (see rows 3 \& 5 in Fig.~\ref{FIG:overallcomparison}).

\subsection{Ablation Study}
We conduct different ablation studies to analyze the proposed method, including the loss functions and our aggregation module. The experiments are evaluated on the DUTS-TEST dataset~\cite{wang2017learning}. We conduct the experiments by combining different parts of our method. As shown in Table~\ref{tbl:ablation}, our method obtains the best performance using all the components, which illustrates that all the loss functions and the AGGM are necessary to realize the one-step training. We use GCPANet~\cite{chen2020global} as our baseline. If the network is directly trained with partial cross entropy loss, the results are relatively low, as listed in 1 of Table~\ref{tbl:ablation}. This phenomenon shows that the partial cross entropy loss is insufficient for sparse labels. Moreover, comparing our final results with the baseline, we obtain gains of 11.61\% for $F_\beta$, 4.73\% for $E_\xi$ and 1.54\% for \emph{MAE}, respectively. 
\begin{table}
	\caption{Ablation study for our losses and AGGM on DUTS-TEST dataset. `Base.' denotes for baseline and `A.' denotes for AGGM. Our overall method obtains the best results.}\label{tbl:ablation}
	\centering
	\begin{tabular}{c|c|c|c|c|ccc}
		\bottomrule
		& Base. & A. & $\mathcal{L}_{ssc}$ & $\mathcal{L}_{lsc}$ & $F_\beta \uparrow$ & $E_\xi \uparrow$ & $MAE\downarrow$ \\
		\midrule
		1 & \checkmark &   &   &   & 0.707 & 0.843 & 0.064 \\
		2 & \checkmark & \checkmark &   &   & 0.706 & 0.845 & 0.064 \\
		3 & \checkmark & \checkmark & \checkmark &   & 0.758 & 0.873 & 0.059 \\
		4 & \checkmark & \checkmark & \checkmark & \checkmark & \textbf{0.823} & \textbf{0.890} & \textbf{0.049} \\
		\bottomrule
	\end{tabular}
\end{table}

\subsubsection{Impact of AGGM.}
\begin{table}
	\caption{Ablation study for our proposed AGGM on DUT-OMRON and DUTS-TEST datasets. It can be seen that our AGGM is compatible to our loss functions.}\label{tbl:aggm}
	\centering
	\begin{tabular}{c|ccc|ccc}
		\bottomrule
		& \multicolumn{3}{| c |}{DUT-OMRON} &\multicolumn{3}{| c }{DUTS-TEST} \\
		& $F_\beta \uparrow$ & $E_\xi \uparrow$ & $MAE\downarrow$ & $F_\beta \uparrow$ & $E_\xi \uparrow$ & $MAE\downarrow$ \\
		\bottomrule
		w/o & 0.730             & 0.845           & 0.069          & 0.800             & 0.877           & 0.053 \\
		w/  & \textbf{0.758}    & \textbf{0.862}  & \textbf{0.060} & \textbf{0.823}    & \textbf{0.890}  & \textbf{0.049} \\
		\bottomrule
	\end{tabular}
\end{table}
In Table~\ref{tbl:aggm}, we evaluate the influence of our aggregation module AGGM when all the loss functions are enabled. It is interesting to see that using AGGM can obtain an average gain of 2.56\% for $F_\beta$, 1.52\% for $E_\xi$ and 0.63\% for \emph{MAE} on DUT-OMRON and DUTS-TEST. However, when training without our proposed losses, as listed in 1 and 2 of Table~\ref{tbl:ablation}, our AGGM contributes little compared with the baseline mode. This phenomenon shows that our AGGM is complementary with the proposed loss functions for sparse labels. 

\subsubsection{Impact of Saliency Structure Consistency Loss.}
\begin{table}
	\caption{Ablation study for SSIM in the saliency structure consistency loss on DUT-OMRON and DUTS-TEST.}\label{tbl:ssim}
	\centering
	\begin{tabular}{c|ccc|ccc}
		\bottomrule
		& \multicolumn{3}{| c |}{DUT-OMRON} &\multicolumn{3}{| c }{DUTS-TEST} \\
		& $F_\beta \uparrow$ & $E_\xi \uparrow$ & $MAE\downarrow$ & $F_\beta \uparrow$ & $E_\xi \uparrow$ & $MAE\downarrow$ \\
		\bottomrule
		w/o    & 0.670             & 0.828           & 0.077          & 0.735             & 0.863           & 0.063 \\
		w/     & \textbf{0.708}    & \textbf{0.841}  & \textbf{0.072} & \textbf{0.758}    & \textbf{0.873}  & \textbf{0.059} \\
		\bottomrule
	\end{tabular}
\end{table}
We conduct this ablation study by adding saliency structure consistency loss to the baseline (AGGM is enabled). The results are shown in 3 of Table~\ref{tbl:ablation}. Compared with only using partial cross entropy loss, using saliency structure consistency loss achieves the improvement of 5.29\% for $F_\beta$, 2.81\% for $E_\xi$ and 0.47\% for \emph{MAE}. That is to say, our saliency structure consistency loss can regularize partial cross entropy loss and enhance the model generalization ability.
Further, we evaluate the impact of SSIM in Eq.~(\ref{eq:structure_consistency_loss}) which is shown in Table~\ref{tbl:ssim}. We train the network using saliency structure consistency loss with and without SSIM separately. It can be seen that the scores of the three evaluation metrics with SSIM are all higher than those without SSIM, which indicates that Eq.~(\ref{eq:structure_consistency_loss}) needs SSIM to make better prediction.

\subsubsection{Impact of Local Saliency Coherence Loss.}
We list the evaluation of the local saliency coherence loss in Table~\ref{tbl:ablation}. With our local saliency coherence loss, the network can obtain the best results. Specifically, using local saliency coherence loss improves $F_\beta$ from 0.7584 to  0.8226, $E_\xi$ from 0.8732 to 0.8904 and \emph{MAE} from 0.0589 to 0.0487. It is the new state-of-the-art performance as reported in Table~\ref{tbl:overall}. Since there is no extra supervision information like edges, such performance demonstrates that our local saliency coherence loss can learn integral salient object structures.  

\section{Conclusions}
In this paper, we have explored one-round training for salient object detection via scribble annotations. We propose a local saliency coherence loss to supervise unlabeled points. Besides, we deploy a self-consistent mechanism via saliency structure consistency loss 
to improve the network generalization ability. Moreover, we have designed an aggregation module to better integrate multiple levels of features, so as to predict better saliency maps for weakly supervised SOD. Experiments show that our approach outperforms previous state-of-the-arts under different evaluation metrics on 6 datasets with a significant margin. Furthermore, our proposed loss functions utilize intrinsic properties of input images to supervise unlabeled points, such that no extra supervision is introduced. 

\section{Acknowledgment}
The work was supported by National Natural Science Foundation of China under 61972323, and Key Program Special Fund in XJTLU under KSF-T-02, KSF-P-02.

\bibliographystyle{aaai21}
\bibliography{reference}
\end{document}